%% file: main.tex
\PassOptionsToPackage{table}{xcolor}
\documentclass[a4paper,fleqn]{cas-dc}

\usepackage{amsmath,amsfonts}
\usepackage{algorithmic}
\usepackage{algorithm}
\usepackage{array}
\usepackage{subfig}
\usepackage{textcomp}
\usepackage{url}
\usepackage{verbatim}
\usepackage{graphicx}
\usepackage{booktabs}
\usepackage{multirow}
\usepackage{tabularx}
\usepackage{float}
\usepackage{rotating}
\usepackage{hyperref}
\usepackage{microtype}
\usepackage[american]{babel}
\usepackage{csquotes}
\providecommand{\IfDocumentMetadataT}[1]{}

\usepackage[backend=biber,style=apa,natbib=true]{biblatex}
\input{apa7-format.tex}
\ExplSyntaxOn
\cs_set:Npn \__first_footerline:
  {
    \group_begin:
    \small\sffamily
    \__short_authors: :~
    {\rmfamily\itshape arXiv~preprint}
    \group_end:
  }
\RenewDocumentCommand \emailauthor { m m }
  {
    \int_gincr:N \g_ead_int
    \seq_gput_right:Nn \g_stm_ead_seq
      {
        \href{mailto:#1}{\nolinkurl{#1}}
        \parsename { #2 }
        \space(\eadauthor)
      }
  }

\ExplSyntaxOff

\definecolor{customcolor}{RGB}{237,4,140}

\hypersetup{
    pdfauthor={Haoran Sun; Chen Cai; Huiping Zhuang; Kong Aik Lee; Lap-Pui Chau; Yi Wang},
    pdfsubject={},
    pdfkeywords={},
    colorlinks=true,
    filecolor=magenta,
    urlcolor=customcolor
}

\begin{document}
\let\WriteBookmarks\relax
\def\floatpagepagefraction{1}
\def\textpagefraction{.001}

\shorttitle{EDVD-LLaMA}
\shortauthors{Sun et al.}

\title[mode=title]{EDVD-LLaMA: Explainable Deepfake Video Detection via Multimodal Large Language Model Reasoning}

\author[1]{Haoran Sun}
\fnmark[1]
\ead{haoran-eee.sun@connect.polyu.hk}
\author[2]{Chen Cai}
\fnmark[1]
\ead{e190210@e.ntu.edu.sg}
\author[3]{Huiping Zhuang}
\ead{hpzhuang@scut.edu.cn}
\author[1]{Kong Aik Lee}
\ead{kong-aik.lee@polyu.edu.hk}
\author[1]{Lap-Pui Chau}
\ead{lap-pui.chau@polyu.edu.hk}
\author[1]{Yi Wang}
\cormark[1]
\ead{yi-eie.wang@polyu.edu.hk}

\affiliation[1]{organization={Department of Electrical and Electronic Engineering, The Hong Kong Polytechnic University},
  city={Hong Kong}, country={China}}
\affiliation[2]{organization={School of Electrical and Electronic Engineering, Nanyang Technological University},
  city={Singapore}, country={Singapore}}
\affiliation[3]{organization={Shien-Ming Wu School of Intelligent Engineering, South China University of Technology},
  city={Guangzhou}, country={China}}
\cortext[1]{Corresponding author.}
\fntext[1]{Haoran Sun and Chen Cai contributed equally to this work.}

\begin{abstract}
The rapid development of deepfake video technology has not only facilitated artistic creation but also made it easier to spread misinformation, which is increasingly difficult to identify. Traditional deepfake video detection (DVD) methods face issues such as a lack of transparency in their principles and insufficient generalization capabilities to cope with evolving forgery techniques. This highlights an urgent need for detectors that can identify forged content and provide verifiable reasoning explanations. This paper proposes the explainable deepfake video detection (EDVD) task and designs the EDVD-LLaMA multimodal, a large language model (MLLM) reasoning framework, which provides traceable reasoning processes alongside accurate detection results and trustworthy explanations. Our approach first incorporates a Spatio-Temporal Subtle Information Tokenization (ST-SIT) to extract and fuse global and local cross-frame deepfake features, providing rich spatio-temporal semantic information input for MLLM reasoning. Second, we construct a Fine-grained Multimodal Chain-of-Thought (Fg-MCoT) mechanism, which introduces facial feature data as hard constraints during the reasoning process to achieve pixel-level spatio-temporal video localization, suppress hallucinated outputs, and enhance the reliability of the chain of thought. In addition, we build an Explainable Reasoning FF++ benchmark dataset (ER-FF++set), leveraging structured data to annotate videos and ensure quality control, thereby supporting dual supervision for reasoning and detection. Extensive experiments demonstrate that EDVD-LLaMA achieves outstanding performance and robustness in terms of detection accuracy, explainability, and its ability to handle cross-forgery methods and cross-dataset scenarios. Compared to previous DVD methods, it provides a more explainable and superior solution.
 The project page is available at: \url{https://11ouo1.github.io/edvd-llama/}.
\end{abstract}


\maketitle

\section{Introduction}
With the increasing popularity of Artificial Intelligence Generated Content (AIGC), deepfake media are easily created using publicly available AIGC video editing tools~\citep{K_2024_CVPR,skorokhodov2022stylegan,yu2023video}.
These techniques can replace faces in videos or alter facial expressions and movements, making such manipulations difficult to detect without a specialized tool \citep{10716705}. While deepfake videos can be used for artistic creation and entertainment, they have serious incidents such as financial fraud, identity theft, and the fabrication of evidence to manipulate public opinion, posing significant information security risks~\citep{hondru2025exddv,shao2022open,li2020celeb}.
Therefore, developing advanced methods for \textbf{D}eepfake \textbf{V}ideo \textbf{D}etection (DVD) has become critically important. In this context, the task of DVD aims to determine whether a video is a deepfake. DVD has broad applications in daily work and life, such as verifying the authenticity of individuals in video calls, filtering out fake videos on social media, and preventing the spread of malicious video content.

\begin{figure*}[pos=!t]
    \centering
    \includegraphics[width=1\textwidth]{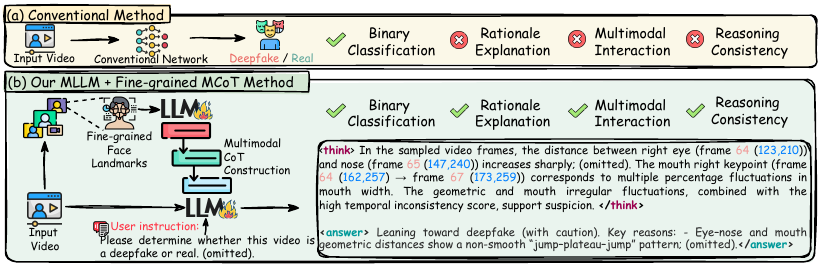}
    \caption{(a) Conventional deepfake video detection: only outputs a binary real/fake label. (b) Our method (MLLM + Fg-MCoT) fuses consecutive frame images and fine-grained facial landmarks to generate a multimodal reasoning chain, enabling a multimodal large language model to achieve multimodal interaction and provide explicit continuous reasoning evidence. ``(omitted)'' indicates content omitted for brevity.}
    \label{fig: Overview}
\end{figure*}

Recent DVD methods have achieved impressive performance by leveraging carefully designed network architectures, sophisticated constraints, and domain-biased pretraining strategies. However, existing DVD approaches face two key challenges that limit their generalization and practicality \citep{Yan_2025_CVPR} (see Fig.~\ref{fig: Overview} (a)). First, most DVD methods are black-box models that only output a binary classification result and a confidence score for a video, without revealing the basis of their decisions to users. Because these methods lack interpretable decision rationales, manual re-examination is still required. Second, in the real world, video deepfake techniques are diverse and continuously evolving; existing DVD methods lack comprehensive generalization ability and cannot ensure satisfactory accuracy when confronted with newly emerging deepfake techniques, thereby significantly undermining their practical value.

\begin{figure}[pos=t]
    \centering
    \includegraphics[width=0.5\textwidth]{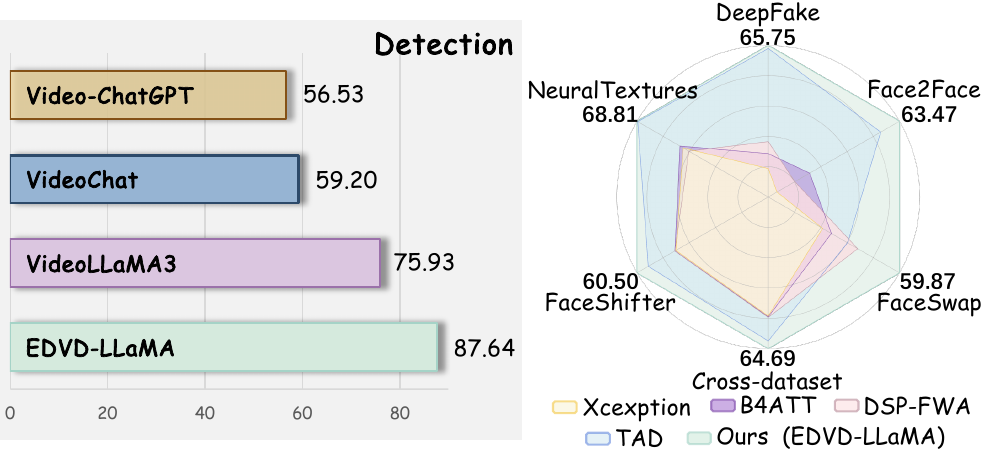}
    \caption{\textit{Left:} Performance comparison between EDVD-LLaMA and MLLMs on the DVD task. \textit{Right:} Performance comparison between EDVD-LLaMA and traditional methods on cross-forgery and cross-dataset detection tasks. EDVD-LLaMA demonstrates superior performance in the above tasks.}
    \label{fig: cross}
\end{figure}

Given that multimodal large language models (MLLMs) are pretrained on large-scale and diverse corpora containing rich world knowledge, recent studies have explored their use for explainable image forgery detection~\citep{zhang2024common,xu2024fakeshield,huang2024ffaamultimodallargelanguage,zhou2025aigi,Yang_2025_CVPR}, demonstrating strong potential for both accurate detection and interpretable explanation. However, these methods share two inherent limitations. First, they operate at the image level and are difficult to extend directly to video, resulting in the loss of temporal information and hindering the recognition of holistic video-level deepfake behaviors. Furthermore, the lack of cross-frame consistency modeling prevents key video deepfake cues, such as boundary artifacts and abnormal physiological rhythms in micro-expressions, from being effectively captured. Second, MLLMs often generate descriptions directly from images, which can lead to hallucinations by introducing forged artifacts that are absent from the input, thereby undermining the credibility of the generated explanations. In addition, although HEIE~\citep{Yang_2025_CVPR} introduces Chain-of-Thought (CoT) reasoning for static image forgery detection and achieves promising results, its reasoning process lacks temporal information and structured forensic evidence as explicit constraints. As a result, its explanations rely largely on free-form textual reasoning, increasing the risk of semantic hallucinations and limiting its generalization to complex and temporally sensitive deepfake video detection.

Motivated by the abovementioned analysis, we propose an \textbf{E}xplainable \textbf{DVD} (EDVD) task and a reasoning framework named EDVD-LLaMA, which is designed to perform deepfake video detection with interpretable reasoning while addressing the two aforementioned limitations.
In the EDVD-LLaMA framework, we first develop a Spatio-Temporal Subtle Information Tokenization (ST-SIT) module to extract and fuse complementary spatio-temporal deepfake features, including fine-grained local artifacts and global semantic anomalies. ST-SIT consists of two complementary branches: a Deepfake Sniffing Encoder (DSEncoder) for extracting local deepfake-sensitive features, and a Compact Visual Connector (CVC) coupled with SigLiP for global feature extraction and spatio-temporal compression. The two branches are subsequently fused to form compact visual representations for LLM reasoning.
Second, we construct a Fine-grained Multimodal Chain-of-Thought (Fg-MCoT) framework based on LLM~\citep{qwen2025qwen25technicalreport}, which incorporates structured facial dynamic metrics as explicit constraints to reduce hallucinations~\citep{10.1162/COLI.a.16} and improve reasoning reliability.
Additionally, we construct an \textbf{E}xplainable \textbf{R}easoning \textbf{FF++} Data\textbf{set} (ER-FF++set) by augmenting existing deepfake video datasets~\citep{roessler2019faceforensicspp} with structured forensic information and leveraging a pretrained MLLM to generate reasoning annotations. Using this dataset, we further fine-tune EDVD-LLaMA for fine-grained reasoning and accurate deepfake video detection.
Our main contributions are summarized as:

\begin{itemize}

\item We present an explainable multimodal deepfake video detection and reasoning framework, termed \textbf{EDVD-LLaMA}. It delivers a trustworthy and comprehensive reasoning trace alongside accurate detection results and their explanations, mitigating the black-box issue inherent in current approaches.

\item We developed ST-SIT to extract and fuse global and inter-frame temporal features for fine-grained spatio-temporal deepfake perception, together with Fg-MCoT that incorporates verifiable facial structured data to reduce hallucinated outputs from the LLM.

\item We utilized a pretrained model to inject verifiable reasoning chains into existing DVD datasets, thereby constructing the ER-FF++set for dual supervision of detection and reasoning. Extensive experiments demonstrate that our method outperforms other MLLMs in deepfake video analysis and detection, while also showing strong generalization across forgery methods and datasets, surpassing most existing DVD methods (Fig.~\ref{fig: cross}).

\end{itemize}

\section{Related Work}
\textbf{Video forgery detection.}
\citet{Yan_2025_CVPR} propose video-level blending data (VB) and spatiotemporal adapters (StA) to improve the generalization of deepfake video detection, and for the first time simulate the ``Facial Feature Drift" forgery artifact. Their method balances spatial and temporal information, but still has room for improvement in handling multimodal forgeries and complex real-world scenarios. The PSO-EfficientNet-GRU method \citep{Cunha_2024} leverages particle swarm optimization in conjunction with deep learning models to significantly enhance the accuracy of video forgery detection. UFCC \citep{electronics13040804} employs content consistency analysis, offering a unified forensic framework that facilitates more stable identification of tampered regions in images and videos. MSIDSnet~\citep{CHENG2024104263} employs multi-scale fusion and a visual Transformer model to achieve robust detection across videos of varying quality. \citet{Siddiqui_2025} combine DenseNet with Cross-ViT to integrate local and global visual features for facial deepfake detection. \citet{Rajesh_2026} extract facial landmarks, texture and shape features, and eye-blink cues, and integrate them through weighted feature fusion and a hybrid 2D--3D convolution-based EfficientNetB7 with attention.

\citet{Khalil_2026} combine spatiotemporal video analysis with LLM-based textual analysis through uncertainty-aware fusion for deepfake detection in asynchronous video--text media. These studies explore complementary visual features, facial dynamics, and multimodal evidence for authenticity classification. Our work further integrates structured facial measurements with local and global video representations to generate evidence-referenced rationales alongside video authenticity decisions.

\textbf{Explainable image forgery detection.}
Traditional interpretable deepfake detection methods primarily construct explainable networks through model design \citep{trinh2021interpretable} or generation operations \citep{shao2022detecting}. For example, DFGNN \citep{Khalid_2023} applies interpretable GNNs to the deepfake detection task. Although these methods enhance the interpretability of the models, they are still unable to describe the reasoning in natural language.
In contrast, recent researchers have utilized MLLMs to construct question-answering tasks~\citep{zhang2024common,huang2024ffaamultimodallargelanguage,xu2024fakeshield}, providing human-understandable explanations for traditional binary classifiers. Recently, CSRDD \citep{zhang2024common} constructs a transferable model based on BLIP, enhancing the generalization ability and interpretability of CNN detectors through multimodal feature fusion. FFAA \citep{huang2024ffaamultimodallargelanguage} redefines deepfake image detection as an open-world visual question answering (VQA) task. They propose a Multi-Answer Intelligent Decision System (MIDS), construct an ``image description–forensics–decision'' triplet dataset, and utilize LLMs combined with multi-hypothesis prompts to generate reasoning explanations for decisions, demonstrating robustness to unknown forgery techniques. FakeShield~\citep{xu2024fakeshield} performs image forgery detection and localization using a LoRA-tuned MLLM that jointly predicts tampering labels, localization rationales, and segmentation masks, improving interpretability and cross-domain generalization. Although these pioneering works achieve interpretable forgery analysis for single images by integrating visual and language reasoning, they are still limited to static images or facial regions.

\textbf{Multimodal large language model.}
Building upon the success of multimodal large language models (MLLMs) in image understanding~\citep{zang2025contextual, cai2025semantics, yin2025rod}, recent studies have extended these models to video understanding by incorporating video encoders and temporal modeling modules to capture both spatial and temporal information~\citep{jin2024chatuniviunifiedvisualrepresentation,liu2023videotellerenhancingcrossmodalgeneration,zhang2023videollamainstructiontunedaudiovisuallanguage}. These approaches typically adopt open-source large language models~\citep{NEURIPS2023_91f18a12,touvron2023llama2openfoundation} as the backbone, with their main differences lying in how video features are encoded, temporally aggregated, and aligned with the language model.
The primary distinction among these works lies in how video features are encoded into visual tokens compatible with LLMs. VideoChat \citep{li2024videochatchatcentricvideounderstanding} employs vision transformers to encode video features and uses a Query Transformer (Q-Former) \citep{li2023blip2bootstrappinglanguageimagepretraining} to compress video tokens. Similarly, VideoLLaMA \citep{zhang2023videollamainstructiontunedaudiovisuallanguage} combines vision transformers (ViT) and image Q-Formers to encode individual frames, followed by temporal modeling using a video Q-Former. While existing multimodal models perform well on short video tasks, such as captioning or question-answering, they lack fine-grained temporal modeling, rendering them ineffective at capturing deepfake traces in long videos.

\textbf{Video question-answering datasets.}
Video question-answering (VQA) is a popular multimodal task, and several large-scale datasets for video-language pretraining have been introduced in recent years. For instance, Ego4D \citep{Grauman_2022_CVPR} and HowTo100M \citep{miech2019howto100mlearningtextvideoembedding} provide massive internet-scale video data for learning video-language understanding. However, the evaluation objectives of these datasets are often narrow, focusing on specific tasks and lacking a comprehensive assessment of the model's fine-grained detection capabilities.

\section{Methodology}

\begin{figure*}[pos=!t]
    \centering
    \includegraphics[width=1\textwidth]{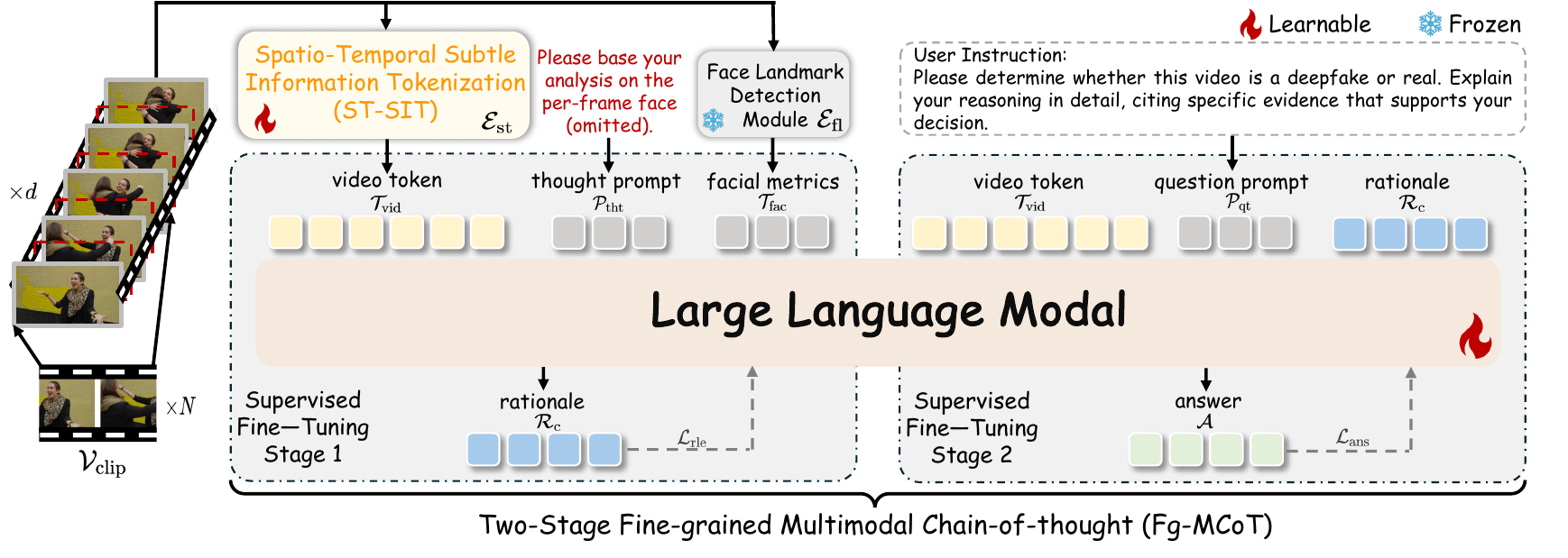}
    \caption{EDVD-LLaMA pipeline. The candidate video is processed by Spatio-Temporal Subtle Information Tokenization (ST-SIT) to obtain video tokens, while the Face Landmark Detection Module provides facial metrics in parallel. In the two-stage Fine-grained Multimodal Chain-of-Thought (Fg-MCoT), Stage 1 combines the video tokens, facial metrics, and thought prompt to generate a rationale. Stage 2 combines the video tokens, rationale, and question prompt to produce the final answer. ``omitted'' indicates content omitted for brevity.}
    \label{fig: pipeline}
\end{figure*}

\subsection{Overview of EDVD-LLaMA Architecture}
Our goal consists of two aspects: (1) leveraging the text comprehension capabilities of MLLM and their prior world knowledge to analyze and establish fundamental principles for determining video authenticity; (2) decoupling principle reasoning from answer inference by constructing Fg-MCoT, thereby reducing hallucination generation in MLLMs and enhancing their ability to make accurate final decisions. To achieve these two tasks, an intuitive approach is to fine-tune an MLLM to output reasoning evidence and final decisions simultaneously. However, we found that jointly training multiple tasks increases the difficulty of network optimization and leads to task interference. Therefore, we propose the EDVD-LLaMA framework, which integrates two key components: the Spatio-Temporal Subtle Information Tokenization (ST-SIT) and the Fine-grained Multimodal Chain-of-Thought (Fg-MCoT) reasoning framework, as illustrated in Fig.~\ref{fig: pipeline}.

Specifically, ST-SIT extracts both global and fine-grained spatio-temporal features from suspicious videos, including local deepfake-sensitive clues and global semantic anomalies. Meanwhile, Fg-MCoT introduces structured facial dynamic metrics as constraints to reduce hallucination generation and improve reasoning reliability. During this process, MLLMs generate detailed reasoning explanations based on these metrics, thereby avoiding domain conflicts caused by diverse forged video features. This design facilitates cross-modal interaction in deepfake detection, enabling MLLMs to achieve trustworthy reasoning and make accurate classifications.

\subsection{Spatio-Temporal Subtle Information Tokenization}
\textbf{Motivation:} In MLLM-based DVD tasks, numerous challenges arise due to the inherent characteristics of video data and the complexity of deepfake techniques. Firstly, video data exhibit strong spatio-temporal dependencies, making it extremely challenging to model long-term sequences and dynamic changes effectively. Secondly, deepfakes often involve subtle spatial manipulations and inter-frame inconsistencies, which current MLLM-based video understanding methods struggle to detect accurately. Additionally, MLLM requires models not only to capture fine-grained visual deepfake cues but also to comprehend high-level semantic information and achieve seamless cross-modal information fusion. Existing methods typically focus on isolated aspects of these issues.
To address these, we propose the Spatio-Temporal Subtle Information Tokenization (ST-SIT) to combine fine-grained spatio-temporal features with global semantic understanding, facilitating a comprehensive analysis of deepfake videos.

We first perform unified video sampling on the input candidate video $\mathcal{V} = \{f_\text{1}, f_\text{2}, ..., f_\text{t}\} \in \mathbb{R}^{t \times H \times W \times 3}$. Using the temporal sampling function, the video is divided into $N$ clips, each containing $d$ frames (in our implementation, $N = 8$, $d = 9$). The sampled video clips are represented as follows:
\begin{equation}
\mathcal{V}_{\text{clip}} = \{ \mathcal{V}_{\text{clip}}^{(i)} \}_{i=1}^{N}, \quad \mathcal{V}_{\text{clip}}^{(i)} = \{ f_{\text{t}}^{(i)} \}_{t=1}^{d}.
\end{equation}
As illustrated in Algorithm \ref{alg:stvfm}, the ST-SIT comprises two feature extraction branches, feature fusion, and token projection. The first branch introduces the DSEncoder based on the Swin Transformer \citep{liu2021swintransformerhierarchicalvision} for local spatio-temporal feature extraction (Lines 1-2). To unify the processing of video frames and explicitly encode temporal information, we propose a 3 × 3 grid layout. Specifically, given a video clip $\mathcal{V}_{\text{clip}}^{(i)} = \{ f_{\text{t}}^{(i)} \}_{t=1}^{d}$, we arrange its resized frames $f_{\text{t}}^{\prime (i)}$ into the 3 × 3 as follows:
\begin{equation}
\mathcal{V}_{\text{grid}}^{(i)} =
\begin{bmatrix}
f_{\text{1}}^{\prime (i)} & f_{\text{2}}^{\prime (i)} & f_{\text{3}}^{\prime (i)} \\
f_{\text{4}}^{\prime (i)} & f_{\text{5}}^{\prime (i)} & f_{\text{6}}^{\prime (i)} \\
f_{\text{7}}^{\prime (i)} & f_{\text{8}}^{\prime (i)} & f_{\text{9}}^{\prime (i)}
\end{bmatrix}, \quad \mathcal{V}_{\text{grid}}^{(i)} \in \mathbb{R}^{3H \times 3W \times 3}.
\end{equation}

\begin{algorithm}[t]
\caption{Spatio-Temporal Subtle Information Tokenization}
\label{alg:stvfm}
\begin{algorithmic}[1]

\REQUIRE Sampled video clip $\mathcal{V}_{\text{clip}}$
\ENSURE Fused spatio-temporal visual representation $\mathcal{T}_{\text{vid}}$

/* Branch 1: Local spatio-temporal feature extraction */
\STATE $\mathcal{V}_{\text{grid}} \leftarrow \mathcal{V}_{\text{clip}}$
\STATE $\mathcal{F}_{\text{ds}} \leftarrow \text{DSEncoder}(\mathcal{V}_{\text{grid}})$

/* Branch 2: Global semantic feature extraction */
\STATE $\mathcal{F}_{\text{sl}} \leftarrow \text{SigLiP}(\mathcal{V}_{\text{clip}})$

\STATE $\mathcal{F}_{\text{cvc}} \leftarrow \text{CVC}(\mathcal{F}_{\text{sl}})$
\STATE \quad \quad $\mathcal{F}_{\text{cmp}} = \text{3DConv}(\mathcal{F}_{\text{sl}})$
\STATE \quad \quad $\mathcal{F}_{\text{cvc}} = \text{RegStage}(\mathcal{F}_{\text{cvc}})$

/* Feature fusion */
\STATE $\mathcal{F}_{\text{fus}} \leftarrow \text{CrossAttention}(\mathcal{F}_{\text{ds}}, \mathcal{F}_{\text{cvc}})$
\STATE \quad \quad $\mathcal{Q}_{\text{ds}} = \mathcal{F}_{\text{ds}} \mathbf{W}_Q,\ \mathcal{K}_{\text{cvc}} = \mathcal{F}_{\text{cvc}} \mathbf{W}_K,\ \mathcal{V}_{\text{cvc}} = \mathcal{F}_{\text{cvc}} \mathbf{W}_V$
\STATE \quad \quad $\text{Attention} = \text{softmax}\left(\frac{\mathcal{Q}_{\text{ds}} \mathcal{K}_{\text{cvc}}^T}{\sqrt{d_k}}\right) \mathcal{V}_{\text{cvc}}$

/* Token projection */
\STATE $\mathcal{F}_{\text{norm}} \leftarrow \text{LayerNorm}(\mathcal{F}_{\text{fus}})$
\STATE $\mathcal{T}_{\text{vid}} \leftarrow \text{Projection}(\mathcal{F}_{\text{norm}})$

\RETURN $\mathcal{T}_{\text{vid}}$

\end{algorithmic}
\end{algorithm}

This grid layout explicitly encodes video temporal dependencies by embedding consecutive frames into spatial neighborhoods. The collection of grid layouts for all video clips is represented as $\mathcal{V}_{\text{grid}} = \{\mathcal{V}_{\text{grid}}^{(1)}, \mathcal{V}_{\text{grid}}^{(2)}, \dots, \mathcal{V}_{\text{grid}}^{(N)}\}$. Using this unified layout as input for the Swin structure~\citep{liu2021swintransformerhierarchicalvision}, the DSEncoder processes consecutive frames as input and employs a sliding window, combined with hierarchical self-attention mechanisms, to extract spatial details and temporal features. DSEncoder explicitly encodes inter-frame information into spatial neighborhoods through the 3 × 3 compact layout. Additionally, it utilizes a multi-stage window extension mechanism, allowing early layers to focus on capturing local deepfake traces and later layers to emphasize global temporal consistency. The entire encoding process progressively downsamples feature maps through hierarchical patch merging, constructing a multi-scale feature hierarchy. This achieves a dynamic balance between computational efficiency and expressive power, generating compact and robust visual embeddings.

The second branch utilizes the SigLip encoder to extract frame-by-frame visual features (Lines 3-6), leveraging its powerful semantic representation capabilities derived from large-scale vision-language pretraining \citep{zhai2023sigmoidlosslanguageimage}. To compress SigLip's frame-by-frame temporal features, we introduce a Compact Visual Connector (CVC). The CVC first utilizes 3D convolution to perform spatio-temporal compression on SigLip's output features, reducing the number of spatio-temporal tokens to accommodate long video inputs. Then, multiple RegStage \citep{radosavovic2020designingnetworkdesignspaces} blocks are used to supplement and enhance spatio-temporal information, thereby avoiding the loss of critical details. With hierarchical stacking and residual connections, CVC ensures gradient flow and training stability, enhancing the model's expressive power and robustness.

The outputs of these two branches, which contain high-dimensional features obtained through deep spatio-temporal modeling and semantic extraction, are integrated via cross-attention \citep{vaswani2017attention} to capture their correlation and complementarity (Lines 7-9). This module effectively combines fine-grained spatio-temporal modeling with semantic understanding, incorporating local and global information to construct fused feature representations. Finally, the fused features are projected to tokens via \citep{liu2023visual} (Lines 10-11) as input to the downstream large language model inference of Fg-MCoT below, enabling precise and robust deepfake detection.

\subsection{Fine-grained Multimodal Chain-of-Thought}
\textbf{Motivation:} Most current CoT frameworks directly feed video encodings into LLMs to perform tasks with intermediate reasoning steps. However, this pipeline often leads to the following issues in DVD tasks: (1) sparse high-frequency deepfake cues, such as fine-grained texture discontinuities, inter-frame lighting variations, and synthesis edge artifacts, are diluted; (2) the lack of a structured ``evidence → reasoning → decision" branch makes it difficult to trace the authenticity of the reasoning process; (3) insufficient modeling of micro-dynamics, such as lip-opening and closing continuity and the smoothness of facial landmark motion. To address these challenges, Fg-MCoT introduces structured facial dynamic metrics that abstract latent deepfake cues from raw pixel dynamics through landmarks and kinematic indicators. These metrics, acting as constraints, are combined with spatio-temporal visual representation from ST-SIT to generate reliable reasoning explanations, suppress hallucinations, and enhance traceability. Ultimately, Fg-MCoT establishes a traceable pipeline, providing a transparent and extensible chain-of-thought framework for DVD tasks, i.e., Explainable DVD (EDVD). Specifically, we propose a systematic task decomposition paradigm in Fg-MCoT, which consists of four coherent stages, progressing from low-level metric cues to high-level semantic decision explanation.

\subsubsection*{Stage 1: Facial Structured Metric Extraction}
We note that focusing the model on the facial features of video characters is crucial for achieving accurate EDVD tasks, as accurately perceiving facial features in the video can make the subsequent reasoning process meaningful. Therefore, our first step is to locate the faces and landmarks of characters in each frame of the video clip, obtaining the basic facial coordinate data to provide traceable ground truth for the subsequent reasoning process. We apply the facial landmark detection $\mathcal{E}_{\text{fl}}$ from open-source dlib\footnote{\url{https://github.com/davisking/dlib}} to extract the facial landmarks:
\begin{equation}
\mathcal{M}_{\text{c}} = \mathcal{E}_{\text{fl}}\{\mathcal{V}_{\text{clip}}\},
\end{equation}
which includes the coordinates of the face region, left and right eyes, nose, and the corners of the mouth, along with a confidence score. Based on the coordinate information, we extract only the largest bounding box in each frame and, following the approach of LipForensics~\citep{haliassos2021lipsdontliegeneralisable}, add 30\% of the face size around the bounding box to refine the bounding box coordinates and write them into $\mathcal{M}_{\text{c}}$. $\mathcal{M}_{\text{c}}$ serves as the supporting preliminary evidence for the next step of facial integrity analysis.

\subsubsection*{Stage 2: Facial Integrity Analysis}We further analyze the corresponding motion behavior by combining the facial coordinates in $\mathcal{M}_{\text{c}}$. In the DVD task, it is insufficient to determine whether there are deepfake features by merely observing static frames. We not only focus on facial pixel information as the primary unit for reasoning, but also on the global perspective of higher-order features of facial landmarks and their interactions. Therefore, to structurally extract potential forgery cues from raw pixel dynamics, we calculate facial integrity metrics $\mathcal{M}_{\Delta}$ based on pixel information across consecutive frames. The specific formulas are as follows.

For the facial region $I_t^{(i)}$ in each frame $f_t^{(i)}$, the variance of the Laplacian transform is calculated to measure the degree of blur:
\begin{equation}
\sigma(I_t^{(i)}) = \frac{1}{N} \sum_{n=1}^{N} \left( \Delta I_{t,n}^{(i)} \right)^2,
\end{equation}
and blur change between consecutive frames is calculated as:
\begin{equation}
\Delta \text{Blur}_{t,t+1}^{(i)} = \left| \sigma(I_t^{(i)}) - \sigma(I_{t+1}^{(i)}) \right|.
\end{equation}
In the LAB color space, the color distribution change between consecutive frames is determined as:
\begin{equation}
\Delta \text{Color}_{t,t+1}^{(i)} = \left| \mu_{t}^{(i)} - \mu_{t+1}^{(i)} \right| + \left| \sigma_{t}^{(i)} - \sigma_{t+1}^{(i)} \right|,
\end{equation}
where \(\mu_{t}^{(i)}\) and \(\sigma_{t}^{(i)}\) are the mean and standard deviation of the facial region in the LAB color space, respectively, calculated:
\begin{equation}
\mu_{t}^{(i)} = \frac{1}{N} \sum_{n=1}^{N} L_{t,n}^{(i)},
\sigma_{t}^{(i)} = \sqrt{\frac{1}{N} \sum_{n=1}^{N} \left( L_{t,n}^{(i)} - \mu_{t}^{(i)} \right)^2}.
\end{equation}
Based on the gray-level co-occurrence matrix (GLCM), the texture contrast is calculated as:
\begin{equation}
\text{Contrast}(G_t^{(i)}) = \sum_{n,m} (n - m)^2 G_{t,n,m}^{(i)},
\end{equation}
and the texture contrast change between consecutive frames is measured as:
\begin{equation}
\Delta \text{Texture}_{t,t+1}^{(i)} = \left| \text{Contrast}(G_t^{(i)}) - \text{Contrast}(G_{t+1}^{(i)}) \right|.
\end{equation} The blending artifacts intensity is calculated based on gradient discontinuity, edge density, and frequency features. Gradient intensity measures the abrupt change in pixel values, calculated as:
\begin{equation}
\text{Gradient}_{t}^{(i)} = \frac{1}{N} \sum_{n=1}^{N} \sqrt{\left( \frac{\partial I_{t,n}^{(i)}}{\partial x} \right)^2 + \left( \frac{\partial I_{t,n}^{(i)}}{\partial y} \right)^2},
\end{equation}
where $\frac{\partial I_{t,n}^{(i)}}{\partial x}$ and $\frac{\partial I_{t,n}^{(i)}}{\partial y}$ are the gradients of pixel $I_{t,n}^{(i)}$ in the $x$ and $y$ directions, respectively. Edge density measures the proportion of edge pixels, expressed as:
\begin{equation}
\text{Edge Density}_{t}^{(i)} = \frac{\text{Number of Edge Pixels in } I_t^{(i)}}{\text{Total Number of Pixels in } I_t^{(i)}}.
\end{equation}
Edge pixels are extracted using the Canny algorithm. Frequency ratio measures the proportion of high-frequency components relative to low-frequency components, defined:
\begin{equation}
\text{Freq Ratio}_{t}^{(i)} = \frac{\sum_{f \in \text{High Freq}} |F^{(i)}(f)|}{\sum_{f \in \text{All Freq}} |F^{(i)}(f)|},
\end{equation}
where $F^{(i)}(f)$ is the discrete Fourier transform (DFT) result of frame $I_t^{(i)}$. Combining the above three metrics, the blending artifact change is computed as:
\begin{align}
\Delta \text{Boundary}_{t,t+1}^{(i)} &=
\left| \text{Gradient}_{t}^{(i)} - \text{Gradient}_{t+1}^{(i)} \right| \nonumber \\
&+ \left| \text{Edge Density}_{t}^{(i)} - \text{Edge Density}_{t+1}^{(i)} \right| \nonumber \\
&+ \left| \text{Freq Ratio}_{t}^{(i)} - \text{Freq Ratio}_{t+1}^{(i)} \right|.
\end{align}
We organize the above metrics into facial integrity metrics $\mathcal{M}_{\Delta}$, including blur variation trends, color distribution consistency, texture feature smoothness, and blending artifacts intensity. $\mathcal{M}_{\Delta}$ is abstracted from raw pixel data, providing clear evidence nodes for downstream rationale generation to suppress hallucinations \citep{10.1162/COLI.a.16}, further constraining the reasoning process to the DVD task. Furthermore, we standardize the data formats of both $\mathcal{M}_{\text{c}}$ and $\mathcal{M}_{\Delta}$ as JSON. This standardized JSON format is designed to facilitate processing by \text{LLM}. $\mathcal{M}_{\text{c}}$ and $\mathcal{M}_{\Delta}$ will serve as fine-grained facial information $\mathcal{T}_{\text{fac}}$ input to \text{LLM}.

\subsubsection*{Stage 3: Rationale Generation}
In this step, as shown in Fig.~\ref{fig: pipeline}, we simultaneously provide $\mathcal{T}_{\text{vid}}$ from ST-SIT, the thought prompt $\mathcal{P}_{\text{tht}}$, and $\mathcal{T}_{\text{fac}}$ from \textit{Stage 2} to $\text{LLM}^{(1)}$ (Supervised Fine-Tuning Stage 1). Notice that the $\mathcal{P}_{\text{tht}}$ is a prompt designed to instruct the model to generate the rationale $\mathcal{R}_{\text{c}}$ (encapsulated in \texttt{<think>}) based on $\mathcal{T}_{\text{vid}}$ and $\mathcal{T}_{\text{fac}}$, such as ``\textit{Please analyze consecutive video frames and the corresponding fine-grained facial information to provide a reasoning process}''. $\text{LLM}^{(1)}$ is capable of generating reliable reasoning that includes traceable and constrained facial information. This contributes to stabilizing the reasoning process of the chain of thought and resisting the interference of hallucination noise. We describe this process as follows:
\begin{equation}
\mathcal{R}_{\text{c}} = \text{LLM}^{(1)} ([\mathcal{T}_{\text{vid}}], [\mathcal{P}_{\text{tht}}], [\mathcal{T}_{\text{fac}}]),
\end{equation}
where $\mathcal{T}_{\text{fac}}$ carries highly concise fine-grained semantic representations, ensuring more precise and reliable tracking of forged features. Moreover, facial keypoint pixel localization based on $\mathcal{T}_{\text{fac}}$ can effectively mitigate the inherent hallucination issues in \text{LLM} \citep{10.1162/COLI.a.16}.

\subsubsection*{Stage 4: Answer Generation}
To encourage \text{LMM} to consider more potential deepfake evidence, we utilize the rationale generated in \textit{Stage 3} as an intermediate step for chain-of-thought reasoning. Consequently, $\text{LLM}^{(2)}$ (Supervised Fine-Tuning Stage 2) combines $\mathcal{T}_{\text{vid}}$, $\mathcal{T}_{\text{fac}}$, and the question prompt $\mathcal{P}_{\text{qt}}$ as the context for providing the final answer $\mathcal{A}$ and reasoning (encapsulated in \texttt{<answer>}). The overall input prompt for answer $\mathcal{A}$ generation is formatted  as:
\begin{equation}
\mathcal{A} = \text{LLM}^{(2)}([\mathcal{T}_{\text{vid}}], [\mathcal{P}_{\text{qt}}], [\mathcal{R}_{\text{c}}]).
\end{equation}

\subsection{Training and Inference}
\subsubsection*{1) Training Objectives}
We adopt a staged optimization strategy that decouples two targets. The first target is learning a metric-grounded rationale. The second target is producing the final authenticity decision. We also enforce alignment between the auxiliary rationale branch and the main decision branch in terms of how they emphasize evidence. $\text{LLM}^{(1)}$ (rationale generator) is first optimized to produce faithful metric-referenced chains \(\mathcal{R}_{\text{c}}\). $\text{LLM}^{(2)}$ (decision head working with ST-SIT) is then trained to output the video level label conditioned on fused visual tokens \(\mathcal{T}_{\text{vid}}\) and the generated rationale. A light joint tuning stage finally introduces an evidence consistency regularizer.

The \(\mathcal{T}_{\text{fac}}\) contains landmark coordinates and derived kinematic and integrity metrics such as blur variance trend, color distribution shift, texture contrast smoothness, and blending artifact indicators and similar descriptors. \(\mathcal{T}_{\text{vid}}\) denotes the fused token sequence output by ST-SIT, which combines the SigLip plus CVC branch with the DSEncoder branch through cross attention. The auxiliary rationale decoder models $p_{\text{asst}}(\mathcal{R}_{\text{c}} \mid \mathcal{T}_{\text{vid}}, \mathcal{T}_{\text{fac}}, \mathcal{P}_{\text{tht}})$,
where \(\mathcal{P}_{\text{tht}}\) is the thought prompt. Its negative log likelihood defines the rationale loss:
\begin{equation}
\mathcal{L}_{\text{rle}} = - \sum_{t=1}^{L_r} \log p_{\text{asst}}(r_t^{*} \mid r_{<t}^{*}, \mathcal{T}_{\text{vid}}, \mathcal{T}_{\text{fac}}, \mathcal{P}_{\text{tht}}).
\end{equation}

The LLM consumes \(\mathcal{T}_{\text{vid}}\), \(\mathcal{R}_{\text{c}}\) and optionally \(\mathcal{T}_{\text{fac}}\) producing a probability \(p_{\text{main}}(y \mid \mathcal{T}_{\text{vid}}, \mathcal{R}_{\text{c}})\) over labels \(y \in \{\text{real}, \text{fake}\}\). The classification loss is:
\begin{equation}
\mathcal{L}_{\text{ans}} = - \log p_{\text{main}}(y^{*} \mid \mathcal{T}_{\text{vid}}, \mathcal{R}_{\text{c}}^{*}).
\end{equation}

To regularize grounding, we construct an evidence candidate set \(\mathcal{E}\) from the keys of \(\mathcal{T}_{\text{fac}}\). For each candidate \(e \in \mathcal{E}\), we derive two distributions. \(p_{\text{asst}}(e \mid \mathcal{T}_{\text{fac}})\): aggregated attention or gating weights over the metric token span when the auxiliary model generates \(\mathcal{R}_{\text{c}}\) followed by temperature scaled softmax. \(p_{\text{main}}(e \mid \mathcal{T}_{\text{vid}}, \mathcal{T}_{\text{txt}})\): analogous attribution distribution when the main model forms its final decision. Here \(\mathcal{T}_{\text{txt}}\) denotes textual inputs including the prompt and injected rationale tokens. We retain a standard KL divergence style penalty on shifts in evidence emphasis:
\begin{equation}
\mathcal{L}_{\text{cons}} = \sum_{e \in \mathcal{E}} p_{\text{asst}}(e\mid \mathcal{T}_{\text{fac}}) \log \frac{p_{\text{asst}}(e\mid \mathcal{T}_{\text{fac}})}{p_{\text{main}}(e\mid \mathcal{T}_{\text{vid}}, \mathcal{T}_{\text{txt}})}.
\end{equation}
The composite loss is:
\begin{equation}
\mathcal{L} = \lambda_1 \mathcal{L}_{\text{rle}} + \lambda_2 \mathcal{L}_{\text{ans}} + \lambda_3 \mathcal{L}_{\text{cons}}.
\end{equation}

\subsubsection*{2) Inference}
Inference follows a modular deterministic pipeline aligned with training. Step 1: Video sampling, uniform or adaptive, produces clips. Step 2: The ST-SIT module is used to yield \(\mathcal{T}_{\text{vid}}\). Step 3: Facial detection and landmark extraction produce \(\mathcal{M}_{\text{c}}\), from which derived integrity and kinematic metrics form \(\mathcal{M}_{\Delta}\). These are serialized into JSON as \(\mathcal{T}_{\text{fac}}\). Step 4: The auxiliary rationale generator produces a metric anchored rationale \(\mathcal{R}_{\text{c}}\). Optionally, multiple rationales can be sampled for uncertainty estimation. Step 5: The LLM consumes \(\mathcal{T}_{\text{vid}}\), \(\mathcal{R}_{\text{c}}\) and traceable \(\mathcal{T}_{\text{fac}}\) and outputs a structured \texttt{<think>} trace and an \texttt{<answer>} label.

\subsection{Construction Process of ER-FF++set}
\textbf{Motivation:} Existing deepfake video datasets only provide binary classification labels for the video modality, lacking visual-language samples suitable for training MLLM in the DVD task. This limitation makes it difficult for MLLM to understand fine-grained instructions related to deepfake details, thereby hindering the development of interpretable reasoning frameworks. Constructing an interpretable reasoning dataset requires leveraging pretrained MLLM to transform the manipulation information in the visual modality of existing DVD datasets into precise textual descriptions. The key challenge in this transformation process lies in suppressing the hallucinated outputs of pretrained MLLM to prevent noise from contaminating the constructed dataset. To address this issue, our core contributions are focused on two aspects: (1) For each deepfake technique, we designed targeted prompts based on its unique characteristics to guide MLLM (like Qwen2.5-VL \citep{Qwen2.5-VL}) in focusing on different deepfake features and providing more accurate visual cues. (2) We provided MLLM with consecutive frames of deepfake videos along with their corresponding masks and facial landmarks, enabling MLLM to generate rationales that focus on manipulated regions, thereby suppressing hallucinated outputs and delivering more accurate deepfake cues.

\begin{figure*}[pos=!t]
    \centering
    \includegraphics[width=1\textwidth]{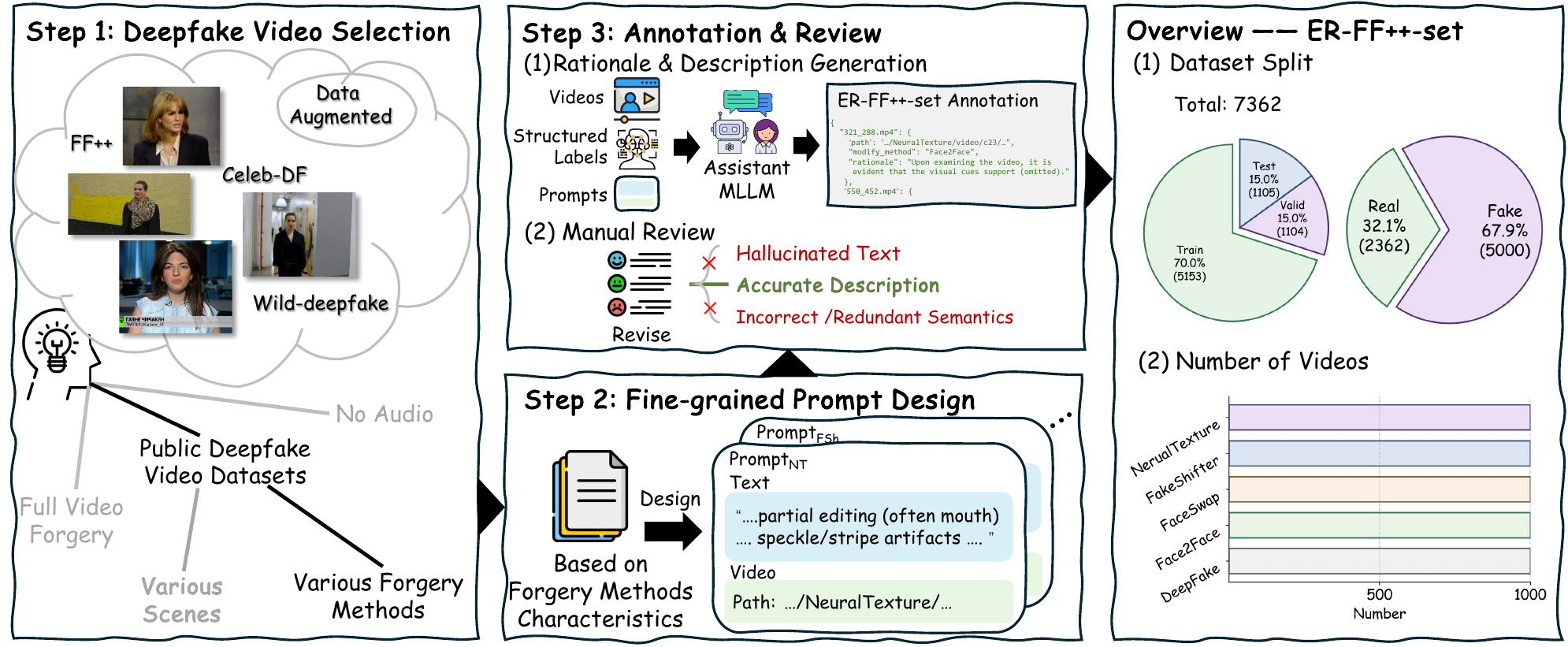}
    \caption{Construction pipeline of the proposed ER-FF++set. We sample videos for multiple manipulation types (Deepfakes, Face2Face, FaceShifter, FaceSwap) from FaceForensics++, and then apply manipulation-type–specific prompts to steer the Assistant LLM in producing structured rationales for each decision.}
    \label{fig: dataset}
\end{figure*}

\textbf{Data Collection:} Based on \citep{roessler2019faceforensicspp}, we categorize common video forgeries into five types: DeepFake \citep{Deepfakes2021}, Face2Face \citep{thies2016face2face}, FaceSwap \citep{FaceSwap2021}, FaceShifter \citep{li2019faceshifter} and NeuralTexture \citep{thies2019deferred}. As shown in step 1 of Fig.~\ref{fig: dataset}, we collect forged videos of these five types, along with their corresponding real videos and self-constructed data, from public datasets.

\textbf{Assistant MLLM-Aided Rationale Generation:} Manual analysis of forged videos is time-consuming and inefficient. Therefore, we employ MLLM~\citep{Qwen2.5-VL} as the assistant LLM for automated analysis of forged videos. The output analysis follows a structured format including detection results, reasoning process, and localization description.

After data collection, to further clarify and concretely describe and analyze the forgery characteristics within videos, the assistant LLM integrates both forgery technique-related features and visible artifacts or semantic errors during annotation generation (as shown in step 2 of Fig.~\ref{fig: dataset}). For visible artifacts and semantic errors, since different manipulation methods produce various types of artifacts, we design dedicated prompts to guide the analysis. These artifacts can be broadly categorized into pixel-level forgery details and image-level semantic errors. For example, NeuralTexture technology often introduces pixel-level issues such as edge artifacts, resolution anomalies, and lighting inconsistencies. Semantic errors are also common, including violations of biological rules or common sense, such as repeated eyebrows on facial regions in DeepFake videos.

In step 3 of Fig.~\ref{fig: dataset}, we first generate forgery region masks \citep{sun2025towards} by pixel-level comparison between real and forged video frames. Subsequently, the video is divided into regions such as the mouth, nose, eyes, and face based on facial landmarks. The degree of forgery in each area is quantitatively assessed to form the structured labels. These structured labels are highly accurate and traceable, effectively reducing hallucinations that may occur during automatic text generation by MLLM. We input the structured labels, region localization information, and dedicated prompts for different forgery techniques into the assistant LLM. For different types of forgery, the LLM conducts targeted analysis based on the structured annotations, quantitatively assesses the degree of forgery in each region, and provides standardized annotations of the tampering method along with rationales that include structured localization information.

The ER-FF++ set contains a total of 7,362 video clips, which are divided according to the proportions shown in the Overview section of Fig.~\ref{fig: dataset}. In terms of authenticity distribution, there are 5,000 Deepfake videos (67.9\%) and 2,362 Real videos (32.1\%). The forgery types cover five mainstream methods (DeepFake \citep{Deepfakes2021}, Face2Face \citep{thies2016face2face}, FaceSwap \citep{FaceSwap2021}, FaceShifter \citep{li2019faceshifter} and NeuralTexture \citep{thies2019deferred}), with each type being evenly distributed. This balanced setting of forgery types and reasonable dataset partitioning help models learn generalizable and interpretable reasoning and detection capabilities under various manipulation mechanisms, and support the evaluation of cross-method generalization and fine-grained reasoning abilities.

\section{Experiments}
\subsection{Experiment Setup}
\textbf{Dataset:} We construct the training and test sets of ER-FF++set using the dataset construction method. For the training set, we use videos manipulated by Deepfake \citep{Deepfakes2021}, Face2Face \citep{thies2016face2face}, FaceSwap \citep{FaceSwap2021}, FaceShifter \citep{li2019faceshifter}, and NeuralTexture \citep{thies2019deferred} techniques, as well as real videos subjected to data augmentation, as the source data. For the test set, in addition to the test portion of ER-FF++set, we also select several challenging public benchmark datasets, including Celeb-DF \citep{li2020celeb} and WildDeepfake \citep{zi2020wilddeepfake}.

\textbf{State-of-the-Art Methods:} To ensure a fair comparison, we selected competitive methods that provide open-source code or pretrained models. To evaluate the DVD performance of EDVD-LLaMA, we compared it with traditional methods such as Xception \citep{chollet2017xception}, DSP-FWA \citep{li2020celeb}, B4 \citep{tan2019efficientnet}, B4ATT \citep{bonettini2021video}, MCX-API \citep{Xu_2023_WACV}, and S-TAD \citep{gao2024texture}. Additionally, we compared it with open-source MLLMs, including Video-LLaVA \citep{lin2023video}, Video-ChatGPT \citep{maaz2024videochatgptdetailedvideounderstanding}, VideoChat \citep{li2024videochatchatcentricvideounderstanding}, and VideoLLaMA3 \citep{zhang2025videollama3frontiermultimodal}. All these methods were retrained on ER-FF++set to ensure consistency with the same training settings. Furthermore, to assess the explainability of EDVD-LLaMA, we compared it with the aforementioned open-source MLLMs (VideoLLaVA \citep{lin2023video}, Video-ChatGPT \citep{maaz2024videochatgptdetailedvideounderstanding}, VideoChat \citep{li2024videochatchatcentricvideounderstanding}, and VideoLLaMA3 \citep{zhang2025videollama3frontiermultimodal}).

\textbf{Evaluation Metrics:} For detection, we report video frame-level accuracy (Acc.), AUC, and F1 score. To evaluate explainability, we adopt multiple automated evaluation metrics, including CIDEr \citep{Vedantam2015CIDEr}, ROUGE\_L \citep{Lin2004ROUGE}, BLEU-4 \citep{Papineni2002BLEU}, METEOR \citep{Denkowski2014METEOR}, and Cosine Semantic Similarity (CSS). These metrics comprehensively measure the similarity between generated and reference texts from various perspectives, including term frequency–inverse document frequency (TF-IDF) weighting, longest common subsequence, n-gram overlap, synonym matching, and semantic vector similarity. For EDVD-LLaMA, the default threshold of 0.5 is applied unless otherwise specified.

\textbf{Implementation Details:} On ER-FF++set, we initially fine-tuned the $\text{LLM}^{(1)}$ (Qwen2.5-7B) \citep{qwen2025qwen25technicalreport} using LoRA (rank=128, alpha=256), while applying full-parameter training to certain components of the ST-SIT module, such as the feature fusion and bridging layers. The training was conducted for 4 epochs on four NVIDIA RTX 6000 Ada 48G GPUs, with a learning rate of 2e-5 and a batch size of 1. Subsequently, we fine-tuned the $\text{LLM}^{(2)}$ (Qwen2.5-7B) \citep{qwen2025qwen25technicalreport} with LoRA using the same configuration, training for 8 epochs on the same hardware setup, with a learning rate of 1e-5 and a batch size of 1. During training, we set the composite loss weights to $\lambda_1=0.8$, $\lambda_2=1$, and $\lambda_3=0.1$ to balance rationale learning, answer prediction, and evidence consistency.

\subsection{Comparison with MLLMs}
To evaluate the deepfake video detection capability of our proposed method, we compare EDVD-LLaMA with several representative MLLMs on the ER-FF++set, as shown in Tab.~\ref{tab:comparison_vllms_DVD}. The results demonstrate that EDVD-LLaMA achieves the best performance across all metrics, with an Acc. of 84.75, AUC of 87.64, and F1 score of 85.13. These results are significantly higher than those of the other MLLMs, such as Video-LLaVA, Video-ChatGPT, and VideoChat, whose accuracy and F1 scores are close to random guessing. Compared to the advanced VideoLLaMA3, our model still surpasses it by a large margin in all metrics (nearly 12\% average improvement). This highlights the superior effectiveness of EDVD-LLaMA for deepfake video detection, benefiting from its tailored feature extraction and interpretable reasoning framework.

\begin{table}[pos=!t]
    \caption{Performance comparison of EDVD-LLaMA with MLLMs on the ER-FF++set for deepfake video detection.}
    \label{tab:comparison_vllms_DVD}
    \centering
    \small
    \resizebox{0.45\textwidth}{!}{
        \begin{tabular}{l|c|ccc}
            \toprule
            \multirow{2}{*}{\textbf{Methods}} & \multirow{2}{*}{\textbf{LLM}} & \multicolumn{3}{c}{\textbf{ER-FF++set}} \\
            \cmidrule(lr){3-5}
             & & Acc.$\uparrow$ & AUC$\uparrow$ & F1$\uparrow$ \\
            \midrule
            Video-LLaVA      & Vicuna-7B & 51.63 & 54.12 & 51.94 \\
            Video-ChatGPT     & LLaVA-7B & 52.27 & 56.53 & 52.38 \\
            VideoChat        & Vicuna-7B & 56.42 & 59.20 & 57.19 \\
            VideoLLaMA3      & Qwen2.5-7B & 72.48 & 75.93 & 73.50 \\
            \rowcolor{gray!25} EDVD-LLaMA (Ours) & Qwen2.5-7B & \textbf{84.75} & \textbf{87.64} & \textbf{85.13} \\
            \bottomrule
        \end{tabular}
    }
\end{table}

To evaluate the quality of explanatory text generation, we compare EDVD-LLaMA with several representative MLLMs on the ER-FF++set using CIDEr, ROUGE\_L, BLEU-4, METEOR, and CSS metrics. As shown in Tab.~\ref{tab:comparison_vllms_text}, EDVD-LLaMA achieves the highest performance across all metrics. Specifically, EDVD-LLaMA obtains a CIDEr score of 2.4579, outperforming VideoLLaMA3 (2.3154), Video-ChatGPT (2.2416), VideoChat (2.1952), and Video-LLaVA (2.1738). Similarly, EDVD-LLaMA achieves the best ROUGE\_L (0.7013), BLEU-4 (0.5048), METEOR (0.4236), and CSS (0.7135), consistently surpassing all baselines.
These results demonstrate that EDVD-LLaMA generates more accurate, fluent, and semantically consistent explanatory texts for deepfake video detection compared to existing MLLMs.
\begin{table}[pos=!t]
    \caption{Performance comparison of MLLMs on the ER-FF++set for explanatory text generation.}
    \label{tab:comparison_vllms_text}
    \centering
    \small
    \resizebox{0.5\textwidth}{!}{
        \begin{tabular}{l|ccccc}
            \toprule
            \multirow{2}{*}{\textbf{Methods}} & \multicolumn{5}{c}{\textbf{Rationale Generation}} \\
            \cmidrule(lr){2-6}
             & CIDEr$\uparrow$ & ROUGE\_L$\uparrow$ & BLEU-4$\uparrow$ & METEOR$\uparrow$ & CSS$\uparrow$ \\
            \midrule
            Video-LLaVA      & 2.1738 & 0.6231 & 0.4165 & 0.3579 & 0.6342 \\
            Video-ChatGPT     & 2.2416 & 0.6542 & 0.4457 & 0.3748 & 0.6573 \\
            VideoChat        & 2.1952 & 0.6318 & 0.4221 & 0.3615 & 0.6457 \\
            VideoLLaMA3      & 2.3154 & 0.6679 & 0.4523 & 0.3842 & 0.6628 \\
            \rowcolor{gray!25} Ours & \textbf{2.4579} & \textbf{0.7013} & \textbf{0.5048} & \textbf{0.4236} & \textbf{0.7135} \\
            \bottomrule
        \end{tabular}
    }
\end{table}

\begin{table}[pos=!t]
\caption{Cross-dataset evaluation results on WildDF and CelebDF. Both MLLM-based and Conventional DVD methods are compared. Cross Avg. denotes the average score across the two public datasets.}
\label{tab:cross-dataset}
\centering
\resizebox{0.48\textwidth}{!}{
\begin{tabular}{l|cc|c}
\toprule
\multirow{2}{*}{\textbf{Methods}} & \multicolumn{2}{c|}{\textbf{Cross Dataset}} & \multirow{2}{*}{\textbf{Cross Avg.}} \\
\cmidrule(lr){2-3}
& WildDF & CelebDF \\
\hline
\multicolumn{4}{l}{\textit{Conventional}} \\
Xception   & 58.03 & 66.35 & 62.19 \\
DSP-FWA    & 53.39 & 57.11 & 55.25 \\
B4         & 63.47 & 73.75 & 68.61 \\
B4ATT      & 62.65 & 69.29 & 65.97 \\
MCX-API    & 50.78 & 56.43 & 53.61 \\
S-TAD        & 64.12 & 64.42 & 64.27 \\
\hline
\multicolumn{4}{l}{\textit{MLLM}} \\
Video-LLaVA    & 50.19 & 53.01 & 51.60 \\
Video-ChatGPT   & 51.66 & 54.27 & 52.97 \\
VideoChat      & 53.78 & 55.84 & 54.81 \\
VideoLLaMA3    & 67.05 & 68.59 & 67.82 \\
\rowcolor{gray!25} Ours           & \textbf{69.11} & \textbf{73.96} & \textbf{71.54} \\
\bottomrule
\end{tabular}}
\end{table}

\subsection{Generalization Performance Cross Forgery Methods and Datasets}

To verify the robustness of the models when confronted with entirely new data distributions, we conducted comprehensive cross-dataset experiments. As presented in Tab.~\ref{tab:cross-dataset}, we evaluated each method on the public WildDF and CelebDF datasets, again comparing both MLLM-based approaches and traditional DVD methods. The experimental results show that our method achieves the highest scores on both datasets, with the cross-dataset average significantly outperforming other approaches, further confirming its excellent generalization performance.

To further evaluate the generalization ability of different methods across various forgery types, we conducted cross-forgery experiments. As shown in Tab.~\ref{tab:cross-forgery}, we performed cross-testing among different forgery types in the ER-FF++set, where models are trained on one type of forgery and tested on the others. The results demonstrate that traditional DVD methods exhibit limited performance when transferring across different types of forgery. In contrast, our proposed method consistently achieves the best average score across forgery methods, indicating superior cross-type generalization capability.

\begin{table}[pos=!t]
\caption{Cross-forgery evaluation results on the ER-FF++set. Models are trained on one type of forgery and tested on others. Cross Avg. denotes the average score across forgery methods.\label{tab:cross-forgery}}
\centering
\resizebox{0.49\textwidth}{!}{
\begin{tabular}{c|c|c|c|c|c|c|c}
\toprule
\multirow{2}{*}{\rotatebox{90}{\textbf{Train}}} & \multirow{2}{*}{\textbf{Methods}}
    & \multicolumn{5}{c|}{\textbf{Test(Intra Dataset——Cross Forgery)}} & \multirow{2}{*}{\begin{tabular}{c}\textbf{Cross} \\ \textbf{Avg.}\end{tabular}}\\
\cmidrule(lr){3-7}
 &  & DF & F2F & FS & FSh & NT & \\
\hline
\multirow{7}{*}{\rotatebox{90}{DeepFake}}
         & Xception & - & 51.25 & 49.87 & 52.03 & 52.92 & 51.52 \\
         & DSP-FWA  & - & 51.73 & 52.03 & 54.94 & 53.17 & 52.97 \\
         & B4       & - & 52.83 & 49.77 & 54.53 & 55.69 & 53.21 \\
         & B4ATT    & - & 52.13 & 49.66 & 52.82 & 54.24 & 52.21 \\
         & MCX-API  & - & 53.34 & 48.82 & 53.61 & 54.56 & 52.58 \\
         & S-TAD    & - & 61.13 & 63.58 & 65.06 & \textbf{70.88} & 65.16 \\
         & Ours     & - & \textbf{63.27} & \textbf{63.72} & \textbf{66.21} & 69.81 & \textbf{65.75} \\
\hline
\multirow{7}{*}{\rotatebox{90}{Face2Face}}
          & Xception & 56.28 & - & 50.73 & 49.74 & 51.43 & 52.05 \\
          & DSP-FWA  & 55.7  & - & 51.63 & 50.76 & 51.8  & 52.47 \\
          & B4       & 58.98 & - & 51.28 & 50.02 & 51.75 & 53.01 \\
          & B4ATT    & 59.71 & - & 51.38 & 49.74 & 51.74 & 53.14 \\
          & MCX-API  & 57.09 & - & 50.76 & 49.89 & 50.76 & 52.13 \\
          & S-TAD    & \textbf{69.6}  & - & 57.24 & 53.42 & 61.41 & 60.42 \\
          & Ours     & 68.51 & - & \textbf{62.39} & \textbf{59.33} & \textbf{63.65} & \textbf{63.47} \\
\hline
\multirow{7}{*}{\rotatebox{90}{FaceSwap}}
         & Xception & 51.22 & 52.67 & - & 50.14 & 49.42 & 50.86 \\
         & DSP-FWA  & 61.13 & 51.52 & - & 52.94 & 50.59 & 54.05 \\
         & B4       & 50.65 & 53.16 & - & 50.27 & 50.09 & 51.04 \\
         & B4ATT    & 51.11 & 54.53 & - & 50.18 & 50.20 & 51.51 \\
         & MCX-API  & 50.44 & 50.33 & - & 49.58 & 48.11 & 49.62 \\
         & S-TAD     & 51.61 & 52.71 & - & 47.90 & 59.61 & 52.96 \\
         & Ours     & \textbf{63.58} & \textbf{58.12} & - & \textbf{57.40} & \textbf{60.37} & \textbf{59.87} \\
\hline
\multirow{7}{*}{\rotatebox{90}{FaceShifter}}
            & Xception & 51.51 & 48.73 & 49.16 & - & 51.18 & 50.15 \\
            & DSP-FWA  & 51.54 & 49.88 & 49.69 & - & 50.34 & 50.36 \\
            & B4       & 50.52 & 49.75 & 49.94 & - & 50.09 & 50.08 \\
            & B4ATT    & 51.23 & 49.71 & 50.17 & - & 50.31 & 50.36 \\
            & MCX-API  & 56.72 & 51.20 & 48.15 & - & 55.01 & 52.77 \\
            & S-TAD     & \textbf{67.99} & 52.68 & 49.43 & - & 58.43 & 57.13 \\
            & Ours     & 66.82 & \textbf{59.07} & \textbf{55.91} & - & \textbf{60.18} & \textbf{60.50} \\
\hline
\multirow{7}{*}{\rotatebox{90}{NeuralTexture}}
               & Xception & 67.06 & 61.21 & 48.46 & 54.81 & - & 57.89 \\
               & DSP-FWA  & 62.36 & 59.18 & 50.18 & 55.79 & - & 56.88 \\
               & B4       & 72.10 & 59.49 & 48.61 & 58.99 & - & 59.8 \\
               & B4ATT    & 67.99 & 60.09 & 48.36 & 57.55 & - & 58.5 \\
               & MCX-API  & 70.88 & 58.98 & 47.41 & 56.64 & - & 58.48 \\
               & S-TAD     & \textbf{82.63} & 63.39 & 62.49 & 65.81 & - & 68.58 \\
               & Ours     & 77.90 & \textbf{65.38} & \textbf{64.51} & \textbf{67.43} & - & \textbf{68.81} \\
\bottomrule
\end{tabular}}
\end{table}

\begin{figure*}[pos=!t]
    \centering
    \includegraphics[width=1\textwidth]{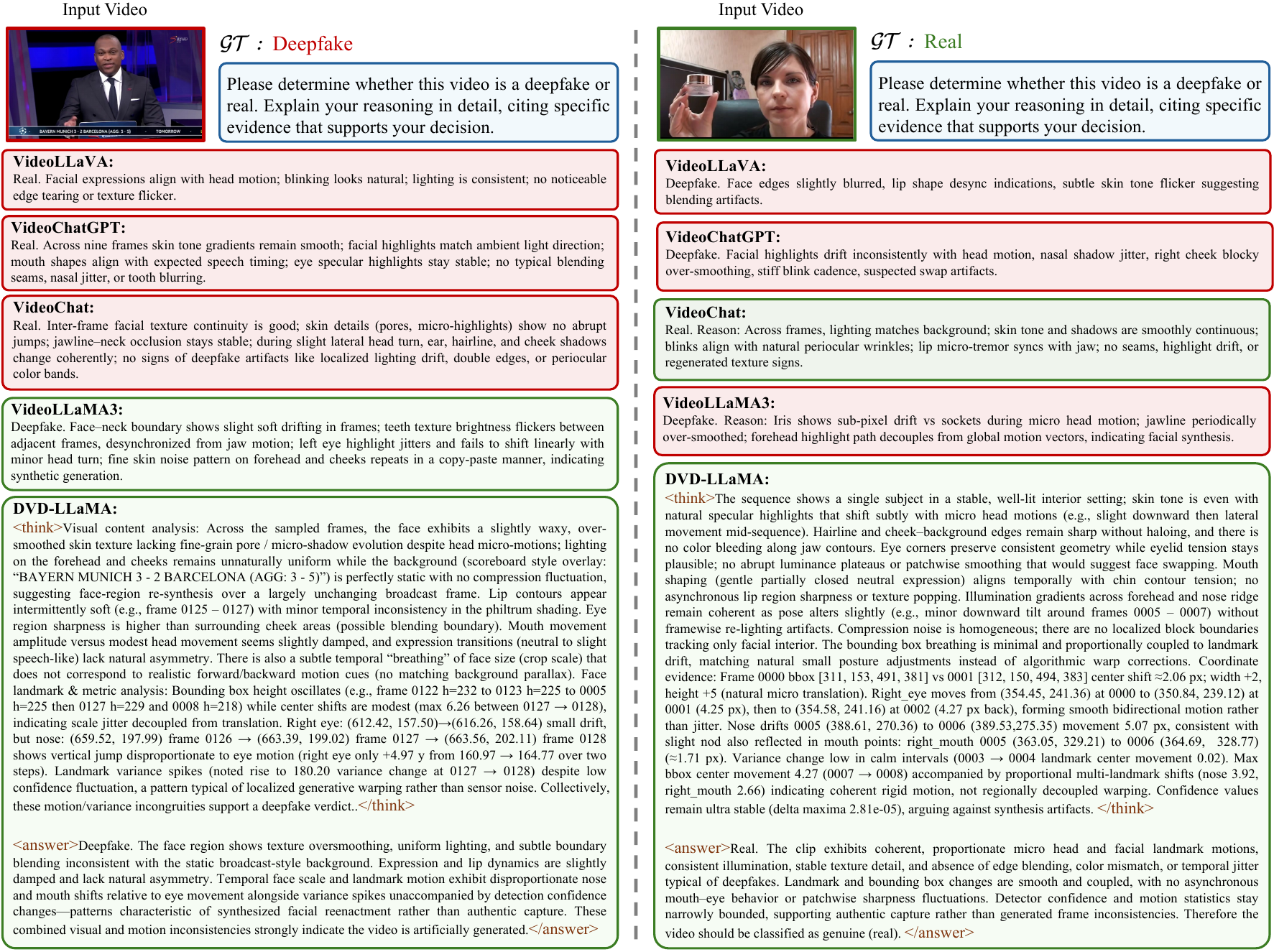}
    \caption{Qualitative results on ER-FF++set. We compare EDVD-LLaMA's test results on deepfake and real videos from ER-FF++ with VideoLLaVA, VideoChatGPT, VideoChat, and VideoLLaMA3. Green boxes indicate correct judgments, and red boxes indicate incorrect decisions. ``omitted'' indicates content omitted for brevity.}
    \label{fig: Qualitative results}
\end{figure*}

\subsection{Qualitative results}
In Fig.~\ref{fig: Qualitative results}, we present a qualitative comparison of EDVD-LLaMA with other baseline models (such as Video-LLaVA \citep{lin2023video}, Video-ChatGPT \citep{maaz2024videochatgptdetailedvideounderstanding}, VideoChat \citep{li2024videochatchatcentricvideounderstanding}, and VideoLLaMA3 \citep{zhang2025videollama3frontiermultimodal}) on the ER-FF++set. This dataset contains samples of deepfake videos and real videos, where the models are required to determine the authenticity of each video. EDVD-LLaMA demonstrates outstanding performance in detecting deepfake videos through its unique design. The results show that EDVD-LLaMA can accurately identify subtle artifacts in deepfake videos, such as decoupled motion vectors and anomalies in dynamic changes. In contrast, other baseline models often fail to handle these complex scenarios due to insufficient sensitivity to artifacts or incomplete feature extraction, resulting in misjudgments. For example, VideoLLaVA and Video-ChatGPT fail to detect critical artifacts in certain deepfake videos, leading to incorrect predictions, while VideoChat and VideoLLaMA3 occasionally generate false positives when processing real videos. In conclusion, EDVD-LLaMA not only achieves accurate judgments of video authenticity but also provides detailed reasoning, proving the effectiveness and potential of its design.

\subsection{Ablation Study}

\textbf{Ablation Study on ST-SIT:}
To validate the importance of the ST-SIT module in the overall performance of EDVD-LLaMA, we designed a series of ablation experiments, progressively removing the key components of ST-SIT, including the DSEncoder ($\mathcal{S}_{\text{dse}}$), CVC ($\mathcal{S}_{\text{cvc}}$), and cross-attention ($\mathcal{A}_{\text{cross}}$). The experimental results are shown in Tab.~\ref{tab: ablation_STVFM}. From the results, it can be observed that when the ST-SIT module is fully retained (\# 1), the model achieves the best results.
When $\mathcal{A}_{\text{cross}}$ (\# 2) is replaced with a linear mapping, the performance drops significantly, with Acc. and F1 decreasing by 3.92 and 3.86, respectively, demonstrating that cross-frame consistency modeling is crucial for capturing temporal features of deepfake video. Furthermore, removing $\mathcal{S}_{\text{cvc}}$ (\# 3) or $\mathcal{S}_{\text{dse}}$ (\# 5) also results in a significant performance drop, further validating the importance of these two components in extracting local deepfake features and fusing cross-frame information. When the ST-SIT module is completely removed (\# 6), the model's performance degrades to baseline levels, with Acc. dropping by 12.27, indicating that the ST-SIT module is a core component for achieving high performance in DVD tasks with EDVD-LLaMA. In summary, ST-SIT effectively extracts and integrates both local and cross-frame spatio-temporal features, providing EDVD-LLaMA with strong forgery perception capabilities and significantly improving detection accuracy.

\begin{table}[pos=!t]
\centering
\caption{Ablation study results on the ER-FF++set. The ST-SIT module makes a significant contribution to the overall performance of EDVD-LLaMA.}
\label{tab: ablation_STVFM}
\resizebox{0.48\textwidth}{!}{
\begin{tabular}{c|ccc|ccc}
\toprule
\# & $\mathcal{S}_{\text{dse}}$ & $\mathcal{S}_{\text{cvc}}$ & $\mathcal{A}_{\text{cross}}$ & Acc. $\uparrow$ & AUC $\uparrow$ & F1 $\uparrow$ \\
\hline
\rowcolor{gray!25} 1 & $\checkmark$ & $\checkmark$ & $\checkmark$ & \textbf{84.75} & \textbf{87.64} & \textbf{85.13} \\
2 & $\checkmark$ & $\checkmark$ & $\times$ & 80.83 & 84.06 & 81.27 \\
3 & $\checkmark$ & $\times$ & $\checkmark$ & 82.98 & 86.17 & 83.44 \\
4 & $\checkmark$ & $\times$ & $\times$ & 76.41 & 79.82 & 76.95 \\
5 & $\times$ & $\checkmark$ & $\times$ & 74.52 & 77.10 & 74.96 \\
6 & $\times$ & $\times$ & $\times$ & 72.48 & 75.93 & 73.50 \\
\bottomrule
\end{tabular}}
\end{table}

\textbf{Ablation Study on Fg-MCoT:}
To verify the effectiveness of the Fg-MCoT, we gradually removed key components from Fg-MCoT, including facial landmarks coordinate information $\mathcal{M}_{\text{c}}$ and facial kinematic data $\mathcal{M}_{\Delta}$ coordinate derivations. The experimental results are shown in Tab.~\ref{tab: ablation_fgmcot}. Retaining Fg-MCoT entirely yields optimal performance on the ER-FF++ set.
When only $\mathcal{M}_{\text{c}}$ is retained and $\mathcal{M}_{\Delta}$ is removed, the performance declines, suggesting that derived computational features contribute to capturing dynamic forgery clues. When both $\mathcal{M}_{\text{c}}$ and $\mathcal{M}_{\Delta}$ are completely removed, the performance further decreases. In this setting, only plain text descriptions are retained without any structured data, which further highlights the necessity of these structured constraints. If Fg-MCoT is entirely removed (w/o Fg-MCoT), the model's performance degrades to an accuracy of 77.15. The above experimental results demonstrate that Fg-MCoT is the core component that enables EDVD-LLaMA to achieve high-precision forgery detection and reliable explainability.

\begin{table}[pos=!t]
\centering
\caption{Ablation study results on Fg-MCoT. Both the $\mathcal{M}_{\text{c}}$ and $\mathcal{M}_{\Delta}$ contribute significantly to the overall performance of EDVD-LLaMA.}
\label{tab: ablation_fgmcot}
\small
\begin{tabular}{l|ccc}
\toprule
Methods & Acc.$\uparrow$ & AUC$\uparrow$ & F1$\uparrow$ \\
\midrule
\rowcolor{gray!25} Ours (w/ Fg-MCoT) & \textbf{84.75} & \textbf{87.64} & \textbf{85.13} \\
\hspace{2.25em} w/\hspace{0.5em} $\mathcal{M}_{\text{c}}$  \& w/o $\mathcal{M}_{\Delta}$ & 82.13 & 86.57 & 82.74 \\
\hspace{2.25em} w/o $\mathcal{M}_{\text{c}}$ \& w/o $\mathcal{M}_{\Delta}$ & 79.04 & 82.91 & 79.42 \\
w/o Fg-MCoT & 77.15 & 80.06 & 77.92 \\
\bottomrule
\end{tabular}
\end{table}

\textbf{Ablation Study on Sampling Strategy: }
To validate the impact of the sampling strategy on performance, we adjusted the number of clips $\mathcal{N}_{\text{c}}$ and the number of frames per clip $\mathcal{P}_{\text{f}}$, and thoroughly evaluated the model performance under the same resolution ($224 \times 224$). The experimental results are shown in Table~\ref{tab: ablation_sampling_strategy}. When the sampling strategy was set to $8 \times 9$, the model achieved the best performance across all metrics. In contrast, reducing $\mathcal{N}_{\text{c}}$ or $\mathcal{P}_{\text{f}}$ resulted in performance degradation, especially when the sampling was set to $4 \times 4$, where the performance dropped significantly. This clearly indicates that fewer clips and frames limit the model's ability to capture holistic forgery cues in videos, thereby reducing the effectiveness of detection.
\begin{table}[pos=!t]
\centering
\caption{Ablation study results on Sampling Strategy. }
\label{tab: ablation_sampling_strategy}
\begin{tabular}{c|c|ccc}
\toprule
$\mathcal{N}_{\text{c}} \times \mathcal{P}_{\text{f}}$ & Size & Acc. & AUC & F1 \\
\hline
\rowcolor{gray!25} $8 \times 9$ & $224 \times 224$ & \textbf{84.75} & \textbf{87.64} & \textbf{85.13} \\
$8 \times 4$ & $224 \times 224$ & 83.42 & 86.23 & 83.81 \\
$4 \times 9$ & $224 \times 224$ & 82.11 & 84.86 & 82.48 \\
$4 \times 4$ & $224 \times 224$ & 78.67 & 81.49 & 79.01 \\
\bottomrule
\end{tabular}

\end{table}

\section{Conclusion}
This paper presents EDVD-LLaMA, a novel framework for explainable deepfake video detection that combines spatio-temporal subtle clue tokenization with fine-grained multimodal reasoning. The proposed approach effectively captures both local and global forgery cues across video frames and utilizes structured facial metrics to enhance the reliability of the reasoning process. The newly constructed ER-FF++set dataset provides comprehensive supervision for both detection and explanation, ensuring high-quality model training. Extensive experimental results demonstrate that EDVD-LLaMA achieves outstanding detection accuracy, robust generalization across different forgery types and datasets, and produces trustworthy, explanatory textual rationales. Ablation studies further confirm the complementary contributions of the feature fusion and reasoning modules. Overall, EDVD-LLaMA establishes a transparent paradigm for deepfake video forensics, offering more reliable and explainable solutions for multimedia security in practical applications.

\section*{Acknowledgments}
The research work was conducted in the JC STEM Lab of Machine Learning and Computer Vision funded by The Hong Kong Jockey Club Charities Trust. This research received partial support from the Global STEM Professorship Scheme from the Hong Kong Special Administrative Region.

\section*{CRediT authorship contribution statement}
Haoran Sun: Methodology, Software, Data curation, Writing - original draft. Chen Cai: Validation, Investigation, Data curation, Writing - original draft. Huiping Zhuang: Writing - review \& editing. Kong Aik Lee: Writing - review \& editing. Lap-Pui Chau: Funding acquisition, Supervision. Yi Wang: Conceptualization, Methodology, Writing - review \& editing, Supervision, Project administration.

\section*{Data availability}
The ER-FF++set reasoning annotations are publicly available at \url{https://huggingface.co/datasets/Codebee/ER-FFppset}. The underlying videos are obtained from the source datasets cited in this paper and remain subject to their respective access conditions and licenses.

\section*{Declaration of generative AI and AI-assisted technologies in the manuscript preparation process}
The authors used OpenAI Codex to check grammar and polish sentence wording. The authors reviewed the revisions and take full responsibility for the manuscript.

\printbibliography[title={References}]

\end{document}

%% file: apa7-format.tex
\DeclareSourcemap{
  \maps[datatype=bibtex]{
    \map{
      \pertype{inproceedings}
      \step[fieldset=month, null]
      \step[fieldset=day, null]
    }
    \map[overwrite]{
      \pertype{misc}
      \step[fieldsource=eprint, final]
      \step[typesource=misc, typetarget=online]
      \step[fieldset=organization, fieldvalue={arXiv}]
    }
    \map{
      \step[fieldsource=eprint, fieldtarget=url, final]
      \step[fieldsource=url, match=\regexp{^(\d{4}\.\d{4,5}(?:v\d+)?)$},
            replace=\regexp{https://arxiv.org/abs/$1}]
    }
    \map{
      \step[fieldsource=howpublished, fieldtarget=url, final]
      \step[fieldsource=url, match=\regexp{\\url\{(.+)\}},
            replace=\regexp{$1}]
    }
    \map[overwrite]{
      \step[fieldsource=title,
            match=\regexp{\b(Cross-ViT|Qwen2\.5(?:-VL)?|[A-Za-z0-9]*[A-Z][A-Za-z0-9]*[A-Z][A-Za-z0-9]*|[A-Z][0-9][A-Za-z0-9]*)\b},
            replace=\regexp{\{$1\}}]
    }
  }
}
\AtEveryBibitem{%
  \ifentrytype{online}{\clearfield{eprint}\clearfield{eprintclass}}{}%
}